# Fare Comparison App of Uber, Ola and Rapido

**Ashlesha Gopinath Sawant, Sahil S. Jadhav, Vidhan R. Jain, Shriraj S. Jagtap, Prachi Jadhav, Soham Jadhav, Ichha Raina**
Computer Science and Engineering (Artificial Intelligence)
Vishwakarma Institute of Technology, Pune, Maharashtra, India

*Abstract:* In today's increasing world, it is very important to have good hailing services like Ola, Uber, and Rapido as it is very essential for our daily transportation. Users often face difficulties in choosing the most appropriate and efficient ride that would lead to both cost-effective and would take us to our destination in less time. This project provides you with the web application that helps you to select the most beneficial ride for you by providing users with the fare comparison between Ola, Uber, Rapido for the destination entered by the user. The backend is use to fetch the data, providing users with the fare comparison for the ride and finally providing with the best option using Python. This research paper also addresses the problem and challenges faced in accessing the data using APIs, Android Studio's emulator, Appium and location comparison. Thus, the aim of the project is to provide transparency to the users in ride-hailing services and increase efficiency and provide users with better experience.

## I. INTRODUCTION

In recent times, the services of the lifts available to us has made our transportation more accessible by offering stronger with several platforms like Ola, Rapido, Uber. These lifts have increased the ease of travelling amongst druggies in India with great vacuity. still, these available platforms also produce a chaos depending upon the factors like business, time of travelling, and also the position. therefore, druggies frequently face difficulties in opting the correct lift option for also that would be more effective for them.

As several platforms are available and each platform having their own fares therefore, for comparing the prices of all of these available platforms, the druggies need to manually check each and every platform for the chow comparison and also elect the effective path. But this is really veritably excited and time consuming for the druggies. This inefficiency highlights the need to have a platform for the druggies where they can check how comparison and elect the lift according to their preferences.

This design aims at erecting a web operation that would give druggies with the price comparison of Ola, Uber and Rapido and also eventually furnishing them with the most suitable option to reach their destination after comparing all the factors. This design provides with real- time data from all the three platforms for the entered volley and drop-off position and also side by side comparing the data.

This paper also includes the specialized frame where all the methodology is mentioned in detail.

The pretensions of this design are
- To produce a web operation that provides the users with the real- time fares of Ola, Uber and Rapido.
- To produce a data for Ola, Rapido and Uber.
- The data of Ola is taken from their system.
- The data of Uber is taken randomly to test the system.
- The data of Rapido is taken from the Bangalore report.

By developing this system, we intend to offer a dependable tool that would help druggies in saving their time and cost at the same time which will enhance translucency amongst druggies.

## II. LITERATURE REVIEW

*Developing and Deploying a Taxi Price Comparison Mobile App in the Wild: Insights and Challenges [1]* The paper discusses the development of the OpenStreetCab which is a mobile application that compares the prices between Uber and (Yellow Cabs in New York and Black Cab in London).This app uses the API to compare the fares of these different platform and also has a enquiry section on their app that help them to take feedback from the users. About 13,000 users installed the app and more than 29000 queries were submitted in the app. After comparing the fares of the platform, it was found that the Uber was expensive than the Yellow Cabs but cheaper than the Black cabs. According to the description, the app shows the relatable prices but the study shows that the app is more efficient in finding time-efficient routes and give options accordingly. This paper showed only 29 journeys considering only two cities for comparison between fares and time

*A Trusted Mobile Ride-Hailing Evaluation System with Privacy and Authentication [2]* The paper mentions the need of privacy for the users when they ride in the cabs available. The ratings given to the driver increases the enhancement for the ride. The rides like Uber and Didi uses kind of cryptographic methods to protect the identity of the driver so that more and more users start trusting on the ride platforms and start taking rides. While selecting the driver, users are able to check the ratings given to the driver and then choose if they want to go with a particular driver.

*Pricing strategy of cab aggregators in India [3]* The ride services in India have been so competent that each platform of transportation is coming up with some new strategies to make their profit. This research paper tells us about different types of strategies that help them to increase their profit. Uber and Ola are competing with each other when it comes to the four-wheeler rides. Rapido has grown a lot as compared to the Ola and Uber when it comes to two-wheelers. The users decide their rides on their priorities of availability, time required to reach their destination and most importantly, the fare comparison.

## III. METHODOLOGY/EXPERIMENTAL

### A. Flowchart & Theory
#### A. Flowchart
Start → Environment Setup → App Installation → Data Extraction → Uber → Ola and Rapido (Automation-based) → End

This project provides you with the web application that helps you to select the most beneficial ride for you by providing users with the real-time fare comparison between Ola, Uber, Rapido for the destination entered by the user. The backend is use to fetch the data and publicly available APIs, providing users with the real-time fare comparison for the ride and finally providing with the best option using Python. For Ola, we have used their general information of prices that is available on their app. For Uber, we have taken a random data to check our system whereas for Rapido, we have considered the Bangalore provided data. This design provides with real-time data from all the three platforms for the entered volley and drop-off position and also side by side comparing the data.

### B. Methodology

Collection of Data
Ola: The data is collected according to the policy of ola. We are following the same method of deciding the fare as per their own system.
Rapido: We have collected Bangalore data for the comparison of fare.
Uber: We have created data on our own for uber services due to difficulties in the accessing APIs.
Accessing data:
For Rapido, we are converting area names into continuous values for calculating the distance between the source and destination.
For Uber, we are using parameters like Distance between source and destination, number of passengers and time of the day.

For Ola, we are accessing the data using the same formula they have used for their app policy.

Comparison of fares:
After fetching the fares of Ola, Uber and Rapido, users are finally able to compare the prices and choose the best possible path.

## C. Testing

Functional Testing
Automation Testing for Ola and Rapido
Error Handling Tests

## IV. RESULTS AND DISCUSSIONS

This app is tested across several routes in an environment emulator and following are the results:
The data fetched from Uber found more accurate. And the data fetched from Ola and Rapido matched with the fares.
All of these available platforms provide users with the ETA (Estimated Time of Arrival) depending upon the traffic condition for those specific platforms.

Conclusion:
The result validates the app in providing real-time fare data and also provided ETA comparison. The web has provided users to find the best option. Although the data of Uber is generated random and not taken from their public APIs is works well and provide results accordingly. One of the important results is that Uber was comparatively faster than the other two. Secondly, after taking recommendation of this app, users save 10-15% on an average after choosing the option shown on the app after comparing prices of all the platforms. Thus, this app help users in their decision making.

## V. FUTURE SCOPE

As this app provides real-time fare comparison of Ola, Uber and Rapido, it is really very beneficial for the users to save their time and money. As it is a good approach towards increasing efficiency, it is very obvious that the future scope of this project is very high.

One of the applications would include directly booking ride through accessing the APIs.

The app for now is only available in urban part, but in future, looking at its efficiency, it might also develop in more and more cities.

Further, some advanced facilities might also available for this app. For example-AC/Non-AC.

Voice command support or a chatbot can also be possible to make this app even for efficient and helpful for users with disabilities.

## VI. CONCLUSION

This project successfully develops an efficient and user-friendly app that helps users to compare the fares of the ride, analyze the trend in price and then selecting the best option amongst Ola, Uber and Rapido to reach to their destination. After combining data of uber taken randomly, Rapido data from available Bangalore report and Data for Ola from their system, this project overcomes all the problems that initially arise due to unavailability of the APIs for Rapido and Ola.
This project not only help users in their transportation but also show users how automation and technology can help us in creating trust within transportation. With proper advancements in this project, it will lead to a more user-friendly platform for the users.

## X. ACKNOWLEDGMENT

We would like to express our gratitude to all those who have supported us throughout this project. We would like to thank our guide, Ms. Ashlesha Gopinath Sawant, for providing valuable guidance throughout this project. We are also grateful and thankful to Mr. C.M. Mahajan, HOD of DESH, and all other faculty members for their immense support during our work. Lastly, we extend our thanks to our team for their cooperation and understanding during project.